%% file: emnlp2020.tex
\documentclass[11pt,a4paper]{article}
\usepackage[hyperref]{emnlp2020}
\usepackage{times}
\usepackage{latexsym}

\usepackage{microtype}

\aclfinalcopy 
\def\aclpaperid{***} 

\title{Linking Entities to Unseen Knowledge Bases with Arbitrary Schemas}

\author{Yogarshi Vyas ~~~ Miguel Ballesteros \\
Amazon AI \\
  \texttt{\{yogarshi, ballemig\}@amazon.com}}

\input{my_latex_preamble}

\date{}

\begin{document}
\maketitle

\begin{abstract}
In entity linking, mentions of named entities in raw text are disambiguated against a knowledge base (KB). This work focuses on linking to unseen KBs that do not have training data and whose schema is unknown during training. Our approach relies on methods to flexibly convert entities from arbitrary KBs with several attribute-value pairs into flat strings, which we use in conjunction with state-of-the-art models for zero-shot linking. To improve the generalization of our model, we use two regularization schemes based on shuffling of entity attributes and handling of unseen attributes. Experiments on English datasets where models are trained on the CoNLL dataset, and tested on the TAC-KBP 2010 dataset show that our models outperform baseline models by over 12 points of accuracy. Unlike prior work, our approach also allows for seamlessly combining multiple training datasets. We test this ability by adding both a completely different dataset (Wikia), as well as increasing amount of training data from the TAC-KBP 2010 training set. Our models perform favorably across the board. 
\end{abstract}

\input{01_intro.tex}
\input{02_related.tex}

\input{03_modeling.tex}
\input{04_data.tex}

\input{05_experiments.tex}

\input{06_conclusion.tex}


\bibliography{anthology,emnlp2020,my_library}
\bibliographystyle{acl_natbib}

\end{document}

%% file: my_latex_preamble.tex
\usepackage{amssymb}
\usepackage{amsthm}
\usepackage{color}
\usepackage{enumitem}
\usepackage{amsmath}
\usepackage{graphicx}
\usepackage{url}
\usepackage{comment}
\usepackage{diagbox}
\usepackage{caption}
\usepackage{verbatim}
\usepackage{mathrsfs}
\usepackage{verbatim}
\usepackage{times}
\usepackage{url}
\usepackage{latexsym}
\usepackage{color}
\usepackage{float}
\usepackage{array}
\usepackage{amsmath,amsfonts,amssymb}
\usepackage{algorithm}
\usepackage{algpseudocode}
\usepackage{graphicx}
\usepackage{booktabs}
\usepackage{pifont}
\usepackage{fancyvrb}
\usepackage{footnote}
\usepackage{array}

\newcolumntype{L}[1]{>{\raggedright\let\newline\\\arraybackslash\hspace{0pt}}m{#1}}
\newcolumntype{C}[1]{>{\centering\let\newline\\\arraybackslash\hspace{0pt}}m{#1}}
\newcolumntype{R}[1]{>{\raggedleft\let\newline\\\arraybackslash\hspace{0pt}}m{#1}}

\usepackage{stmaryrd}

\def\Argmax{\operatornamewithlimits{arg\,max}\limits}

\newcommand{\mnote}[1]
{
}

\newcommand{\eg}{{\it e.g.}}
\newcommand{\viz}{{\it viz.}}

\newcommand{\ie}{{\it i.e.}}

\newcommand{\ensuretext}[1]{#1}
\newcommand{\yvmarker}{\ensuretext{\textcolor{green}{\ensuremath{^{\textsc{Y}}_{\textsc{V}}}}}}
\newcommand{\mbmarker}{\ensuretext{\textcolor{blue}{\ensuremath{^{\textsc{M}}_{\textsc{B}}}}}}

\newcommand{\mycomment}[3]{}
\newcommand{\yv}[1]{\mycomment{\yvmarker}{#1}{green}}
\newcommand{\mb}[1]{\mycomment{\mbmarker}{#1}{blue}}

\newcommand{\cell}[2]{{#1} {\scriptsize $\pm$ #2}}

\setlist{noitemsep}
\newcommand{\ignore}[1]{}

%% file: 01_intro.tex
\section{Introduction}

Entity linking consists of linking mentions of entities found in text against canonical entities found in a target \textit{knowledge base} (KB). Early work in this area was motivated by the availability of large scale knowledge bases containing millions of entities~\cite{BunescuPasca2006}. A large fraction of subsequent work has followed in this tradition of linking to a handful of large, publicly available KBs such as Wikipedia, DBPedia~\cite{AuerBizerKobilarovLehmannCyganiakIves2007} or the KBs used in the now decade-old TAC-KBP challenges~\cite{McnameeDang2009,JiGrishmanDangGriffittEllis2010}. As a result, previous work always assumes complete knowledge of the \textit{schema} of the target KB that entity linking models are trained for, \ie{} how many and which \textit{attributes} are used to represent entities in the KB. This allows training supervised machine learning models that exploit the schema along with labeled data that link mentions to this \textit{a priori} known KB. This strong assumption, however, breaks down in many scenarios which require linking to KBs that are not known at training time. For example, a company might want to automatically link mentions of its products to an internal KB of products that has a rich schema with several attributes, \eg{} product category, description, dimensions, etc. It is very unlikely that the company will have training data of this nature, \ie{} mentions of products linked to its database.

Our focus is on this problem of linking entities to unseen KBs with arbitrary schemas. One solution is to annotate data that can be used to train specialized models for each target KB of interest, but this is not scalable. A more generic solution is to build entity linking models that can work with arbitrary KBs. We follow this latter approach and build entity linking models that can link to completely unseen target KBs that have not been observed during training.\footnote{``Unseen KBs" refers to scenarios where we neither know the entities in the KB, nor its schema.}

Our solution builds on recently introduced models for zero-shot entity linking~\cite{WuPetroniJosifoskiRiedelZettlemoyer2020,LogeswaranChangLeeToutanovaDevlinLee2019}. However, these models assume the same, simple schema during training and inference. Instead, we generalize these models and allow them to handle arbitrary (and different) KBs during training and inference, containing entities represented with an arbitrary set of \textit{attribute-value} pairs.

This generalization relies on two key ideas. First, we use a series of methods to convert arbitrary entities (from any KB), into a string representation that can be consumed by the models for zero-shot linking. Central to the string representation are special tokens called \textbf{attribute separators}, which are used to represent frequently occurring attributes in the training KB(s), and carry over their knowledge to unseen KBs during inference (Section~\ref{subsec:att2text}). Second, we generate more flexible string representations by shuffling entity attributes before converting them to strings, and  by stochastically removing attribute separators to improve generalization to unseen attributes (Section~\ref{subsec:reg}).

Our primary experiments are cross-KB and focus on English datasets. We train models to link to one dataset during training (\viz{} Wikidata), and evaluate them for their ability to link to an unseen KB (\viz{} the TAC-KBP Knowledge Base). These experiments reveal that our model with \textbf{attribute-separators} and the two generalization schemes are 12--14 points more accurate than the baseline zero-shot models used in an \textit{ad hoc} way\ignore{ (\eg{}, by just pasting together all attributes separated by the [SEP] tokens from BERT~\cite{DevlinChangLeeToutanova2019})}. Ablation studies reveal that while all model components individually contribute to this improvement, combining all of them results in the most accurate models. 

Finally, unlike previous work, our models allow seamless mixing of multiple training datasets which link to potentially different KBs with different schemas. We investigate the impact of training on multiple datasets \mb{in the abstract you mention CoNLL and TAC-KBP only, rephrase that} in two sets of complementary experiments involving additional training data that a) links to a third KB that is different from our original training and testing KBs, and b) links to the same KB as the test KBs. These experiments reveal that our models perform favorably under all  conditions compared to the baselines.

%% file: 02_related.tex
\begin{table*}[!t]
\scalebox{0.85}{
    \centering
    \begin{tabular}{l|cccc}
    \toprule
       & \textbf{Generic EL} & \textbf{Zero-shot EL} & \textbf{Linking to any DB} & \textbf{This work} \\
       & & \cite{LogeswaranChangLeeToutanovaDevlinLee2019} & \cite{SilCroninNieYangPopescuYates2012} & \\
       \midrule

       Test entities seen during training & Yes & No & No & No\\
       Test KB schema known  & Yes & Yes & Yes & No \\
       In-domain test data & Yes & No & Yes & Not necessarily\\ 
       Restricted Candidate Set & No & No & Yes & No \\ 
       \bottomrule
    \end{tabular}
    }
    \caption{This table compares the entity linking framework in the present work with those in previous work.}
    \label{tab:compare}
\end{table*}

\section{Background}

Conventional entity linking focuses on settings where models are trained on the KB that they are evaluated on \cite{BunescuPasca2006}. Typically, this KB is either Wikipedia, or derived from Wikipedia in some way \cite{LingSinghWeld2015}. This limited scope allows models to avail of other sources of information to improve linking, including (but not limited to) alias tables, frequency statistics, and rich metadata.

\paragraph{Beyond Conventional Entity Linking} There have been several attempts to move beyond such conventional settings, such as by moving beyond Wikipedia to KBs from diverse domains such as the biomedical sciences \citep{ZhengHowsmonZhangHahnMcGuinnessHendlerJi2014,DSouzaNg2015} and music~\citep{OramasAnkeSordoSaggionSerra2016} or even being completely domain and language independent~\citep{WangZhengMaFoxJi2015,OnoeDurrett2020}.  \citet{LinLinJi2017} discuss approaches to link entities to a KB that simply contains a list of names without any other information. \citet{SilCroninNieYangPopescuYates2012} perform linking against database using database-agnostic features. However, their approach still requires training data from the target KB. \ignore{In contrast, this work aims to train entity linking models that do not rely on training data from the target KB, and can be trained on arbitrary KBs, and applied to a different set of KBs. \citet{} also attempt to link entities from any domain and language but their core idea is fundamentally opposite to our work in that they explicitly aim to model the structure of a particular target KB (DBPedia), while we aim to be KB-agnostic.} \citet{PanCassidyHermjakobJiKnight2015} also do \textit{unsupervised} entity linking by generating rich context representations for mentions using Abstract Meaning Representations \cite{BanarescuBonialCaiGeorgescuGriffittHermjakobKnightKoehnPalmerSchneider2013}, followed by unsupervised graph inference to compare contexts. More recently, \citet{LogeswaranChangLeeToutanovaDevlinLee2019} have introduced a novel zero-shot entity linking framework to ``develop entity linking systems that can generalize to unseen specialized entities".  Table \ref{tab:compare} summarizes how the entity linking framework considered in this work differs from a few of these works.

\paragraph{Contextualized Representations for Entity Linking} Models in this work are based on BERT, a pre-trained language model for  contextualized representations that has been successfully used for a wide range of NLP tasks \cite{DevlinChangLeeToutanova2019}. While many studies have tried to understand why BERT performs so well~\citep{RogersKovalevaRumshisky2020}, the work by \citet{TenneyDasPavlick2019} is most relevant as they use probing tasks to show that BERT encodes knowledge of entities. This has also been shown empirically by many works that use BERT and other contextualized models for entity linking and disambiguation \citep{Broscheit2019,ShahbaziFernGhaeiniObeidatTadepalli2019,YamadaWashioShindoMatsumoto2020,FevryFitzGeraldKwiatkowski2020,PoernerWaltingerSchutze2020}.

 \ignore{However, the schema of the target KB remains constant across model training and evaluation. Specifically, \citet{LogeswaranChangLeeToutanovaDevlinLee2019} assume that the entities are linked to an entity dictionary, which consists of the name of the entity paired with a textual description of the entity~(\eg, the first few lines of the Wikipedia page of that entity). \ignore{While the present work uses models originally developed for this task~\cite{WuPetroniJosifoskiRiedelZettlemoyer2019}, it makes no such assumptions about the schema and instead provides a solution that can handle unseen KBs.}}

%% file: 03_modeling.tex
\section{Preliminaries}
\label{sec:bg}
\subsection{Entity Linking Setup}

Entity linking consists of disambiguating entity mentions $\mathcal{M}$ from one or more documents to a target knowledge base, $\mathcal{KB}$, containing unique entities. We assume that each entity $e \in \mathcal{KB}$ is represented using a set of attribute-value pairs $\{(k_i,v_i)\}_{i=1}^{n}$. The attributes $k_i$ collectively form the \textit{schema} of $\mathcal{KB}$. The disambiguation of each $m \in \mathcal{M}$ is aided by the \textit{context} $c$ in which $m$ appears. 

Models for entity linking typically consist of two stages that balance recall and precision. 

\begin{enumerate}
\item \textbf{Candidate generation}:  The objective of this stage is to select $K$ candidate entities $\mathcal{E} \subset \mathcal{KB}$ for each mention $m \in \mathcal{M}$, where $K$ is a hyperparameter and $K << |\mathcal{KB}|$. Typically, models for candidate generation are less complex (and hence, less precise) than those used in the following (re-ranking) stage since they handle all entities in $\mathcal{KB}$. Instead, the goal of these models is to produce a small but high-recall candidate list $\mathcal{E}$. Ergo, the success of this stage is measured using a metric such as recall@$K$ \ie{} whether the candidate list contains the correct entity. \\

\item \textbf{Candidate Reranking}: This stage ranks the candidates in $\mathcal{E}$ by how likely they are to be the correct entity. Unlike candidate generation, models for re-ranking are typically more complex and oriented towards generating a high-precision ranked list since the objective of this stage is to identify the most likely entity for each mention. This stage is evaluated using precision@1 (or accuracy) \ie{} whether the highest ranked entity is the correct entity.
\end{enumerate}

In a traditional entity linking setup, the training mentions $\mathcal{M}_\textit{train}$ and test mentions $\mathcal{M}_\textit{test}$ both link to the same KB. Even in the zero-shot settings of \citet{LogeswaranChangLeeToutanovaDevlinLee2019}, while the training and target domains (and hence the knowledge bases) are mutually exclusive, the schema of the KB is constant and known. On the contrary, our goal is to link test mentions $\mathcal{M}_\textit{test}$ to a knowledge base $\mathcal{KB}_\textit{test}$ which is not known during model training. \ignore{Thus, not only do we not know the kind of entities that are covered by $\mathcal{M}_\textit{test}$, we also do not know the schema of $\mathcal{KB}_\textit{test}$.} The objective is to train models on mentions $\mathcal{M}_\textit{train}$ that link to $\mathcal{KB}_\textit{train}$ and directly use these models to link $\mathcal{M}_\textit{test}$ to $\mathcal{KB}_\textit{test}$.

\subsection{Zero-shot Entity Linking}

The starting point (and baselines) for our work are the state-of-the-art models for zero-shot entity linking, which we briefly describe here~\cite{WuPetroniJosifoskiRiedelZettlemoyer2020,LogeswaranChangLeeToutanovaDevlinLee2019}.\footnote{We re-implemented these models and verified them by comparing results with those in the original papers.}

\paragraph{Candidate Generation} Our baseline candidate generation approach relies on similarities between mentions and candidates in a vector space to identify the candidates for each mention \citep{WuPetroniJosifoskiRiedelZettlemoyer2020} using two BERT models \cite{DevlinChangLeeToutanova2019}. The first BERT model encodes a mention $m$ along with its context $c$ into a vector representation $\mathbf{v}_\textit{m}$. $\mathbf{v}_\textit{m}$ is obtained from the reserved [CLS] token used in BERT models to indicate the start of a sequence. In this encoder, mention words are additionally indicated by a special embedding vector that is added to the token embeddings of the mention as in \citet{LogeswaranChangLeeToutanovaDevlinLee2019}. The second (unmodified) BERT model independently encodes each $e \in \mathcal{KB}$ into vectors. The candidates $\mathcal{E}$ for a mention are the $K$ entities whose representations are most similar to $\mathbf{v}_\textit{m}$. Both BERT models are fine-tuned jointly using a cross-entropy loss to maximize the similarity between a mention and its corresponding correct entity, when compared to other random entities.

\paragraph{Candidate Re-ranking} The candidate re-ranking approach uses a BERT-based cross-attention encoder to jointly encode a mention and its context along with each candidate from $\mathcal{E}$~\cite{LogeswaranChangLeeToutanovaDevlinLee2019}. Specifically, the mention $m$ is concatenated with its context on the left ($c_l$), its context on the right ($c_r$), and a single candidate entity $e \in \mathcal{E}$. An [SEP] token, which is used in BERT to separate inputs from different segments, is used here to separate the mention in context, from the candidate. This concatenated string is encoded using BERT (again, with the special mention embeddings added to the mention token embeddings) to obtain, $\mathbf{h}_{m,e}$ a representation for this mention/candidate pair (from the [CLS] token). Given a candidate list $\mathcal{E}$ of size $K$ generated in the previous stage, $K$ scores are generated for each mention, which are subsequently scored using a dot-product with a learned weight vector ($\mathbf{w}$). Thus,
\begin{gather*}
     \mathbf{h}_{m,e} = \text{BERT}(\text{[CLS]}~c_l~m~c_r~\text{[SEP]}~e~\text{[SEP]}), \\
     \textit{score}_{m,e} = \mathbf{w}^{T}\mathbf{h}_{m,e}.
 \end{gather*}
 The candidate with the highest score is chosen as the correct entity, \ie{}
\begin{gather*}
     e^* = \Argmax_{i=1}^{K}~~\textit{score}_{m,e_i}.
\end{gather*}

\section{Linking to Unseen Knowledge Bases}
\label{sec:modeling}

The models in Section~\ref{sec:bg} were designed to operate in settings where the entities in the target KB were only represented using a textual description. For example, the entity \textit{Douglas Adams} would be represented in such a database using a description as follows:

{\textit{Douglas Noel Adams was an English author, screenwriter, essayist, humorist, satirist and dramatist. Adams was author of The Hitchhiker's Guide to the Galaxy.}}

However, linking to unseen KBs requires handling entities that have an arbitrary number and type of attributes. The same entity (\textit{Douglas Adams}) can be represented in a different KB using attributes such as ``name", ``place of birth", etc. as shown at the top of Figure~\ref{fig:sep}. This raises the question of whether such models, that harness the power of large-scale, pre-trained language models, generalize to linking mentions to unseen, including those where such textual descriptions are not available. In this section, we present multiple ideas to this end.

\subsection{Representing Arbitrary Entities using Attribute Separators}
\label{subsec:att2text}

One way of using these models for linking against arbitrary KBs is by defining an \textit{attribute-to-text} function $f$, that maps arbitrary entities with any set of attributes $\{k_i,v_i\}_{i=1}^{n}$ to a string representation $e$ that can be consumed by BERT, \ie{} 
\begin{gather*}
    e = f(\{k_i,v_i\}_{i=1}^{n}).
\end{gather*}
If all entities in the KB are represented using such string representations, then the models described in Section~\ref{sec:bg} can directly be used for arbitrary schemas. This then leads to a follow-up question: \textit{how can we generate string representations for entities from arbitrary KBs such that they can be used for BERT-based models}? Alternatively, what form can the function $f$ take?

An obvious answer to this question is simple \textbf{concatenation} of all the values of an entity, given by 
\begin{gather*}
    f(\{k_i,v_i\}_{i=1}^{n}) = v_1~v_2~...~v_n.
\end{gather*}

We can improve on this by adding some structure to this representation by teaching our model that the $v_i$ belong to different segments. As in the baseline candidate re-ranking model, we do this by separating them with [SEP] tokens. We call this \textbf{[SEP]-separation}.
\begin{gather*}
    f(\{k_i,v_i\}_{i=1}^{n}) = \text{[SEP]}~v_1~\text{[SEP]}~v_2~...~\text{[SEP]}~v_n
\end{gather*}

\begin{figure}[t]
    \centering
    \includegraphics[trim={2.8in 4in 1.9in 0},clip,width=\linewidth]{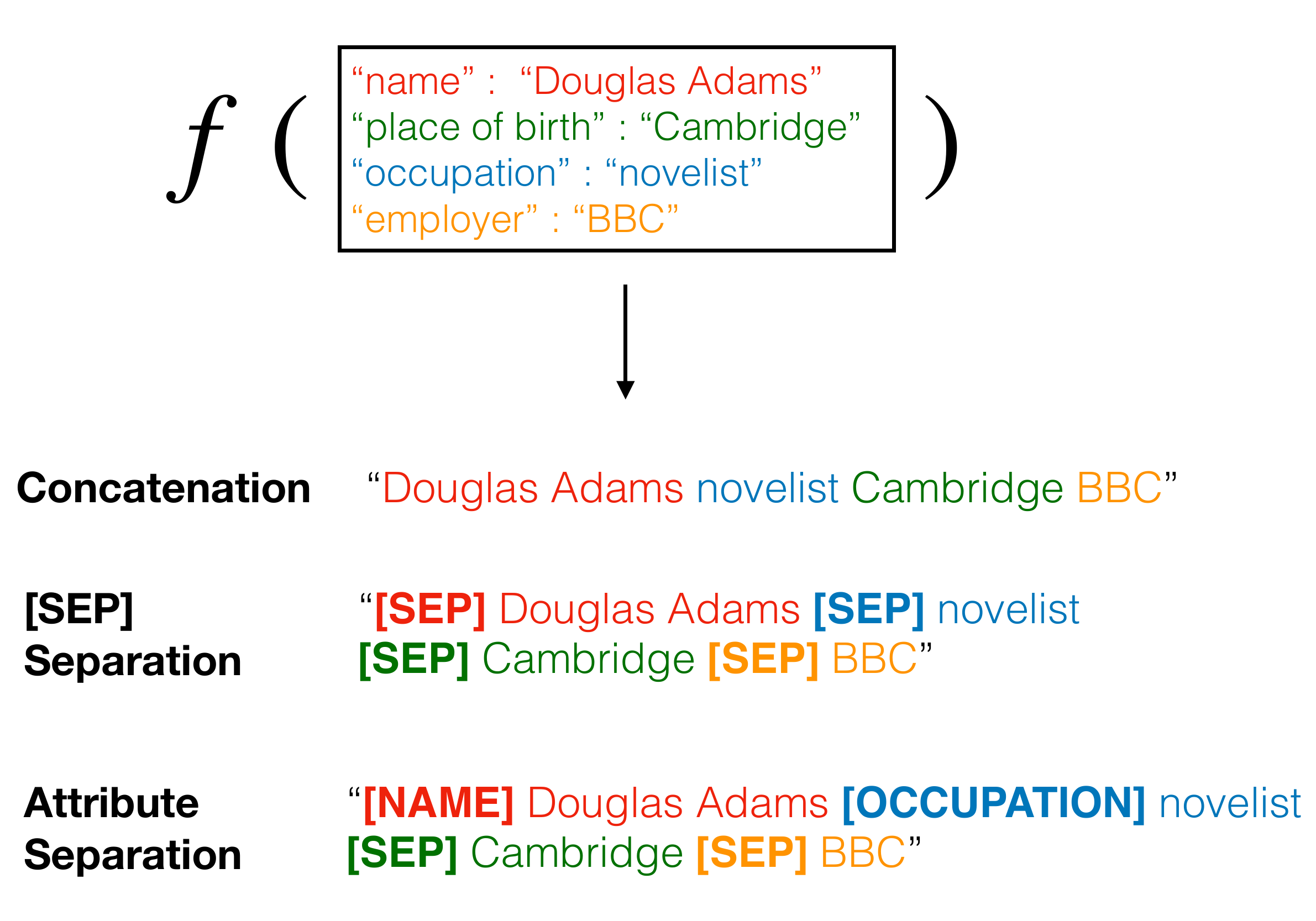}
    \caption{Shown here are the three ways of representing an entity with arbitrary attribute-values (Section~\ref{subsec:att2text}). \textbf{Concatenation} simply concatenates all values, \textbf{[SEP]-separation} separates attributes using [SEP] tokens, and \textbf{attribute separation} introduces special tokens based on the most frequently occurring attributes in the training data (which in this toy example are ``name" and ``occupation").}
    \label{fig:sep}
\end{figure}

In the above two definitions of $f$, we have used the values $v_i$ of the entity, but not the attributes $k_i$, which also contain meaningful information. For example, if an entity seen during inference has a \textit{capital} attribute with the value ``\textit{New Delhi}'', seeing the \textit{capital} attribute allows us to infer that the target entity is likely to be a place, rather than a person, especially if we have seen the \textit{capital} attribute during training. We use this information in the form of \textbf{attribute separators}, which are reserved tokens (in the vein of [SEP] tokens) that correspond to attributes. In this case, 
\begin{gather*}
    f(\{k_i,v_i\}_{i=1}^{n}) = \text{[$K_1$]}~v_1~\text{[$K_2$]}~v_2~...~\text{[$K_n$]}~v_n.
\end{gather*}

These [$K_i$] tokens are not part of the vocabulary of the BERT model, so they do not have pre-trained embeddings as other tokens in the vocabulary. Instead, we augment the existing vocabulary with these new tokens and introduce them during training the entity linking model(s) based on the most frequent attribute values seen in the target KB of the training data, and randomly initialize their corresponding embeddings. During inference, when we are faced with an unseen KB, we use attribute separators for only those attributes that we have seen during training, and choose to use the [SEP] token for the remaining attributes.

Figure~\ref{fig:sep} illustrates the three different instantiations of $f$. In all cases, attribute-value pairs are ordered in descending order of the frequency with which they appear in the training KB. Finally, since both the candidate generation and candidate re-ranking models we build on (Section~\ref{sec:bg}) use BERT, the techniques discussed here can be applied to both stages, but we only focus on the re-ranking stage. We defer more details to Section~\ref{sec:setup}.

\subsection{Regularization Schemes for Improving Generalization}
\label{subsec:reg}

Building  models for entity linking against unseen KBs requires that such models do not overfit to the training data by memorizing characteristics of the training KB. This is done by using two regularization schemes that we apply on top of the candidate string generation techniques discussed in the previous section.

\paragraph{Attribute-OOV}
The first scheme, which we call \textbf{attribute-OOV}, prevents models from overtly relying on individual [$K_i$] tokens and generalize to attributes that are not seen during training. Analogous to how out-of-vocabulary tokens are commonly handled \cite[\textit{inter alia}]{DyerBallesterosLingMatthewsSmith2015}, we stochastically replace every [$K_i$] token during training with a [SEP] token with probability $p_{\textit{drop}}$. This encourages the model to encode semantics of the attributes in not only the individual [$K_i$] tokens, but also in the [SEP] token, which is then used when unseen attributes are encountered during inference.

\paragraph{Attribute-shuffle}
The second regularization scheme discourages the model from memorizing the order in which particular attributes occur. Under \textbf{attribute-shuffle}, every time an entity is encountered during training, its attribute/values are randomly shuffled before they are converted to a string representation using any of the techniques described in Section~\ref{subsec:att2text}.

%% file: 04_data.tex
\section{Experimental Setup}
\label{sec:setup}

\subsection{Data}
Our held-out test bed is the TAC-KBP 2010 data which consists of documents from English newswire, discussion forum and web data ~\cite{JiGrishmanDangGriffittEllis2010}.\footnote{\url{https://catalog.ldc.upenn.edu/LDC2018T16}} The target KB, $\mathcal{KB}_\textit{test}$, is the TAC-KBP Reference Knowledge Base which is built from English Wikipedia articles and their associated infoboxes.\footnote{\url{https://catalog.ldc.upenn.edu/LDC2014T16}} Our primary training and validation data is the CoNLL-YAGO dataset~\cite{HoffartYosefBordinoFurstenauPinkalSpaniolTanevaThaterWeikum2011}, which consists of documents from the CoNLL 2003 Named Entity Recognition task \cite{TjongKimSangDeMeulder2003} linked to multiple KBs including Wikipedia.\footnote{\url{https://www.mpi-inf.mpg.de/departments/databases-and-information-systems/research/yago-naga/aida/downloads/}} To ensure that our training and target KBs are different, we use \textit{Wikidata} as our training KB.\footnote{Retrieved from \url{https://dumps.wikimedia.org/wikidatawiki/entities/} in March, 2020.} Specifically, we use the subset of entities from Wikidata that have a Wikipedia page. We ignore all mentions that do not have a corresponding entity in the KB, both during training and inference, leaving the task of handling such NIL entities to future work. Finally, we use the Wikia dataset from \citet{LogeswaranChangLeeToutanovaDevlinLee2019} for experiments with investigate the impact of multiple datasets (Section~\ref{subsec:multidataset}).\footnote{\url{https://github.com/lajanugen/zeshel}}
Table~\ref{tab:datastats} summarizes these datasets.

\begin{table}[t]
    \centering
    \begin{tabular}{lrr}
    \toprule
     & Number of & Size of\\
    & mentions & target KB\\
    \midrule
    CoNLL-YAGO (train) & 18527 & 5.7M\\
    CoNLL-YAGO (val.) & 4788 & 5.7M\\
    Wikia (train) & 49275 &  0.5M\\
    Wikia (val.) & 10000 & 0.5M\\
    TAC KBP 2010 (test) & 1658 & 0.8M\\
    \bottomrule
    \end{tabular}
    \caption{Number of mentions in our training, validation, and test sets, along with the number of entities in their respective KBs.}
    \label{tab:datastats}
\end{table}

\yv{Explain difference between two KBs and why it matters for us}
While covering similar domains, Wikidata and the TAC-KBP Reference KB have a few significant differences that make them suitable for our experiments. First, and most relevant to this work, they have highly different schemas. Wikidata is more structured and entities are associated with statements represented using attribute-value pairs, which are typically short snippets of information rather than full sentences. On the other hand, the TAC-KBP Reference KB contains both short snippets like these, along with the entire textual contents of the Wikipedia article corresponding to the entity. The two KBs differ in size, with Wikidata containing almost seven times the number of entities in TAC KBP.

Both during training and inference, we only retain the 100 most frequent attributes in the respective KBs. The attribute-separators described in Section~\ref{subsec:att2text} are created corresponding to the 100 most frequent attributes in the training KB. The embeddings for these tokens are randomly initialized using a Gaussian distribution with zero mean and unit variance.

\subsection{Training details and hyperparameters}

\label{subsec:config}

All BERT models are uncased BERT-base models with 12 layers, 768 hidden units, and 12 heads with default parameters, and trained on English Wikipedia and the BookCorpus.  The probability $p_\textit{drop}$ for \textbf{attribute-OOV} is set to 0.3.

Both candidate generation and re-ranking models are trained using the BERT Adam optimizer~\cite{KingmaBa2014}, with a linear warmup for 10\% of the first epoch to a peak learning rate of $2\times10^{-5}$ and a linear decay from there till the learning rate approaches zero.\footnote{\url{https://gluon-nlp.mxnet.io/api/modules/optimizer.html\#gluonnlp.optimizer.BERTAdam}} Candidate generation models are trained for 200 epochs with a batch size of 256. Re-ranking models are trained for 4 epochs with a batch size of 2, and operate on the top 32 candidates returned by the generation model. Candidates and mentions (with context) are represented using strings of 128 sub-word tokens each, across all models.  Hyperparameters are chosen such that models can be run on a single NVIDIA V100 Tensor Core GPU with 32 GB RAM, and are not extensively tuned. All re-ranking experiments are run with five different random seeds, and we report the mean and standard deviation of the accuracy across all runs.

%% file: 05_experiments.tex
\section{Experiments and Discussion}

We evaluate the accuracy of the re-ranking architecture from Section~\ref{sec:bg} under different conditions, using a fixed candidate generation model. We aim to answer the following research questions:

\begin{enumerate}[noitemsep]
    \item Do the attribute-to-text functions (Section~\ref{subsec:att2text}) generate useful string representations for arbitrary entities? Specifically, can these string representations be used in concordance with the re-ranking model from Section~\ref{sec:bg} to link to the unseen $\mathcal{KB}_\textit{test}$ ?
    \item How much impact do the three key components of our model --- \textbf{attribute-separators} (Section~\ref{subsec:att2text}), \textbf{attribute-shuffling}, and \textbf{attribute-OOV} (Section~\ref{subsec:reg}) --- individually have on our model?
    \item Does training on more than one KB with different schemas help models in more accurately linking to $\mathcal{KB}_\textit{test}$?
    \item Do improvements for generalizing to unseen $\mathcal{KB}_\textit{test}$ also translate to improvements in scenarios where there is training data that also links to $\mathcal{KB}_\textit{test}$?
\end{enumerate}

\subsection{Candidate Generation Results} 

Before we focus on our research questions, we briefly discuss our candidate generation model. Since the focus of our experiments is primarily on re-ranking, we do not extensively experiment with the candidate generation model, and use a single model that combines the architecture of \citet{WuPetroniJosifoskiRiedelZettlemoyer2020} (Section~\ref{sec:bg}) with \textbf{[SEP]-separation} to generate candidate strings. This model is trained on the CoNLL-Wikidata dataset, and achieves a recall@32 of 91.25 when evaluated on the TAC-KBP 2010 set\mb{we should mention that this is held-out evalutation, we do not use TAC-KBP recall@32 as stopping criteria}. This model also has no knowledge of the schema of the KB seen during inference. \ignore{In contrast, models that are schema-aware have achieved recall@32 of yy \cite{}. We leave it to future work to close this gap.  }\yv{Need to find what is SOTA recall for this dataset..}\mb{maybe you can just train your model on TAC-KBP, dev TAC-KBP and test TAC-KBP}

\subsection{Main results} 
\label{subsec:mainres}

\begin{table}[t]
    \centering
    \begin{tabular}{lr}
    \toprule
        Model & Accuracy \\
        \midrule
        \textbf{concatenation}  & \cell{47.2}{7.9} \\
        \textbf{[SEP]-separation}  & \cell{49.1}{2.6} \\
        \textbf{attribute-separation} (no reg.) & \cell{54.7}{3.8}  \\
        ~~~++\textbf{attribute-OOV} & \cell{56.2}{2.5} \\
        ~~~++\textbf{attribute-shuffle} & \cell{58.2}{3.6}\\
        ~~~++\textbf{attribute-OOV} + {shuffle} & \cell{61.6}{3.6} \\ 
        \midrule
        \midrule
        \citet{RaimanRaiman2018} & 90.9 \\
        \citet{CaoHouLiLiu2018} & 91.0 \\
        \citet{WuPetroniJosifoskiRiedelZettlemoyer2020} & 94.0 \\
        \citet{FevryFitzGeraldKwiatkowski2020} & 94.9 \\
    \bottomrule
    \end{tabular}
    \caption{Training on CoNLL-Wikidata and testing on the TAC-KBP 2010 test set reveals that using \textbf{attribute-separators} instead of [SEP] tokens yields string representations for candidates that result in more accurate models. Regularization schemes (Section~\ref{subsec:reg}) further improve accuracy to 61.6\% on the TAC-KBP 2010 test set without using any training data from that KB. }
    \label{tab:exp1}
\end{table}

In our primary experiments, we focus on the first two research questions and study the accuracy of the model that uses the re-ranking architecture from Section~\ref{sec:bg} with the three core components introduced in Section~\ref{sec:modeling} \viz{} \textbf{attribute-separators} to generate string representations of candidates, along with \textbf{attribute-OOV} and \textbf{attribute-shuffle} for regularization. We compare this against two baselines without these components that use the same architecture and use \textbf{concatenation} and \textbf{[SEP]-separation} instead of \textbf{attribute-separators}.\footnote{The baselines have the same parameters as our models with attribute separators, except that the latter have 100 extra token embeddings (of size 768 each) for the attribute-separators.} As a reminder, all models are trained as well as validated on CoNLL-Wikidata and evaluated on the completely unseen TAC-KBP 2010 test set. 

Results (Table~\ref{tab:exp1}) confirm that adding structure to the candidate string representations in the form of [SEP] tokens leads to more accurate models compared to generating strings by concatenation. We also observe that using \textbf{attribute-separators} instead of [SEP] tokens leads to a gain of over 5 accuracy points. Using \textbf{attribute-OOV} to handle unseen attributes further increases the accuracy to 56.2\%, a 7.1\% increase over the [SEP] baseline. Taken together, these results demonstrate the use of \textbf{attribute-separators} in capturing meaningful information about attributes, even when only a small number of attributes from the training data (15) are observed during inference.

Shuffling attribute-value pairs before converting them to a string representation using \textbf{attribute-separators} also independently provides an accuracy gain of 3.5 points over the model which uses \textbf{attribute-separators} without shuffling. Overall, combining \textbf{attribute-shuffling} and \textbf{attribute-OOV} yields  the most accurate models  with an accuracy of 61.6, which represents a 12 point accuracy gain over the best baseline model. 

The most accurate results in Table~\ref{tab:exp1} are still over 30 points behind the state-of-the-art models on this dataset~\cite{RaimanRaiman2018,CaoHouLiLiu2018,WuPetroniJosifoskiRiedelZettlemoyer2020,FevryFitzGeraldKwiatkowski2020}. However, there are three key differences between our models and the most accurate models. First, state-of-the-art models are completely supervised in that they use in-KB training data. On the contrary, the purpose of this work is to show how far we can go without using such in-KB data. Second, these models always rely only on the textual description of the entity in the KB. On the contrary, our models are not trained on the test KB, and can flexibly work with arbitrary schemas that have a diverse set of attributes. Finally, beyond the in-KB data, these models are also pre-trained on the entirety of Wikipedia for the task of linking (which amounts to 17M training mentions in the case of ~\citet{FevryFitzGeraldKwiatkowski2020}). On the other hand, the focus of this work is on establishing the effectiveness of linking to unseen KBs and we leave it to future work to close the gap by using such pre-training.

\subsection{Training on multiple unrelated datasets}
\label{subsec:multidataset}

\begin{table}[t]
    \centering
    \begin{tabular}{lr}
    \toprule
        Model & Accuracy \\
        \midrule
        \textbf{[SEP]-separation}  & \cell{62.6}{0.8} \\
        \textbf{attribute-separation} &  \\
        ~~~++\textbf{attribute-OOV} + \textbf{shuffle} & \cell{66.8}{2.8} \\ 
    \bottomrule
    \end{tabular}
    \caption{Adding the Wikia dataset to training improves accuracy of both our model and the baseline, but our models still outpeform the baseline by over 4 points.}
    \label{tab:exp2datasets}
\end{table}

An additional benefit of being able to link to multiple KBs is the ability to train on more than one datasets, each of which can link to a different KB with different schemas. While prior work has been unable to do so due to its reliance on knowledge of  $\mathcal{KB}_\textit{test}$, this ability is more crucial in the settings investigated in this work, as it allows us to stack independent datasets for training. This allows us to answer our third research question. Specifically, we compare the \textbf{[SEP]-separation} baseline with our full model that uses \textbf{attribute-separators}, \textbf{attribute-shuffle}, and \textbf{attribute-OOV}. We ask whether the differences observed in Table~\ref{tab:exp1} also hold when these models are trained on a combination of two datasets \viz{} the CoNLL-Wikidata and the Wikia datasets, before being tested on the TAC-KBP 2010 test set.

Adding the Wikia dataset to the training increases the accuracy of the full model by 6 points, from 61.6 to 66.8 (Table~\ref{tab:exp2datasets}). In contrast, the baseline model observes a bigger increase in accuracy from 49.1 to 62.6. While the difference between the two models reduces, our full model still remains more accurate. These results also show that the seamless stacking of more than one dataset allowed by our models is also effective empirically.

\subsection{Impact of schema-aware training data}
\label{subsec:schemaexp}

Finally, we turn to our fourth and final question and investigate to what extent do components introduced in this work help in linking when there is training data available that links to the inference KB, $\mathcal{KB}_\textit{test}$. We hypothesize that while \textbf{attribute-separators} will still be useful, \textbf{attribute-OOV} and \textbf{attribute-shuffle} will be less useful as there is a smaller gap between training and test scenarios, reducing the need for regularization. 

For these experiments, models from Section~\ref{subsec:mainres} are further trained with data from the TAC-KBP 2010 training set. A sample of 200 documents is held out from training data to use as a validation set. To observe model behavior in different data conditions, we run these next set of experiments with 1\%, 5\%, 10\%, 25\%, 50\%, 75\%, and 100\% of the available training data.\footnote{The 0\% results are the same as those in Table~\ref{tab:exp1}.}  For simplicity, these samples are obtained at the document level, and not the mention level. Thus, since the TAC training data has 1300 documents, 1\% corresponds to 13 documents, and so on.  The models are trained with the exact same configuration as the base models (Section~\ref{subsec:config}), except using a constant learning rate of $2 \times 10^{-6}$.

\begin{table}[t]
    \centering
    \scalebox{0.95}{
    \begin{tabular}{lrrr}
    \toprule
        \% of TAC  & \textbf{[SEP]-sep.} & \multicolumn{2}{c}{\textbf{Attribute-sep.}} \\
        training data & & w/ reg. & w/o reg.\\
        \midrule
        0\% & \cell{49.1}{2.6} & \multicolumn{2}{c}{\cell{61.6}{3.6}} \\
        1\% & \cell{62.4}{3.1}	& \cell{69.0}{0.5} & \cell{70.0}{2.8} \\
        5\% & \cell{70.1}{2.5}	& \cell{72.8}{1.5} & \cell{76.0}{1.6} \\
        10\% & \cell{74.5}{2.0}	& \cell{76.0}{0.8} & \cell{77.8}{1.6} \\
        25\% & \cell{80.1}{1.2}	& \cell{78.8}{0.4} & \cell{80.8}{1.0} \\
        50\% & \cell{81.8}{1.0}	& \cell{80.5}{0.4} & \cell{82.8}{1.1} \\
        75\% & \cell{83.1}{1.0}	& \cell{81.1}{0.2} & \cell{84.0}{0.5}  \\
        100\% & \cell{84.1}{0.6}	& \cell{81.8}{0.9} & \cell{84.9}{0.7}  \\
        \midrule
        \midrule
        TAC-only &  	& \cell{83.6}{0.7} & \cell{83.8}{0.9}  \\
    \bottomrule
    \end{tabular}
    }
    \caption{Experiments with increasing amounts of training data that links to the inference KB reveal that models with \textbf{attribute separators} but without any regularization are the most accurate across the spectrum.}
    \label{tab:expdata}
\end{table}

Perhaps unsurprisingly, accuracy of all models increases as the amount of TAC training data increases (Table~\ref{tab:expdata}). Also, as hypothesized, the smaller generalization gap between training and test scenarios makes the model with only \textbf{attribute separators} more accurate than the model with both \textbf{attribute separators} and regularization. 

Crucially, however, the model with only \textbf{attribute separators} is consistently the most accurate model across the spectrum of additional data. Moreover, the difference between this model and the baseline model sharply increases as the amount of schema-aware data decreases.  In fact, just by using 13 annotated documents (\ie{} 1\% of the training data), we get a 9 point boost in accuracy over the completely zero-shot model. These trends shows the models in this work are not only useful in settings without any data from the target KB, but also in those where very limited data is available.

While the last two rows in Table~\ref{tab:expdata} now observe the same in-KB training data as the state-of-the-art models in Table~\ref{tab:exp1}, the differences highlighted in Section~\ref{subsec:mainres} still remain --- models in Table~\ref{tab:expdata} are still not pre-trained on millions of mentions in Wikipedia, and these models can still be flexibly used with unseen KBs as they are not optimized for the TAC-KBP dataset.

%% file: 06_conclusion.tex
\section{Conclusion}

The primary contribution of this work is in introducing a novel setup for entity linking against unseen target KBs with unknown schemas. To this end, we introduce methods to generalize existing models for zero-shot entity linking to link to arbitrary KBs during both training and inference. These methods rely on converting arbitrary entities represented using a set of attribute-value pairs into a string representation that can be then consumed by models from prior work.

As results indicate, there is still a significant gap between schema-aware models that are trained on the same KB as the inference KB, and models used in this work. One way to close this gap could be by using automatic table-to-text generation techniques to convert arbitrary entities into fluent and adequate text \cite{Kukich1983,McKeown1985,ReiterDale1997,WisemanShieberRush2017,ChisholmRadfordHachey2017}. Another promising direction is to move beyond BERT to other pre-trained representations that are better known to encode entity information~\cite{ZhangHanLiuJiangSunLiu2019,GuuLeeTungPasupatChang2020,PoernerWaltingerSchutze2020}. 

Finally, while the focus of this work is only on English entity linking, challenges associated with this work naturally occur in multilingual settings as well. Just as we cannot expect labeled data for every target KB of interest, we also cannot expect labeled data for different KBs in different languages. In future work, we aim to investigate how we can port the solutions introduced here to multilingual settings as well develop novel solutions for scenarios where either the documents or the KB (or both) are in languages other than English \cite{Sil2018NeuralCE,Upadhyay2018JointMS}.